%% file: cvpr.tex
\documentclass[final]{cvpr}

\usepackage{times}
\usepackage{epsfig}
\usepackage{graphicx}
\usepackage{amsmath}
\usepackage{amssymb}
\usepackage{comment}
\usepackage{stfloats}

\usepackage{algorithm}
\usepackage{algpseudocode}
\algtext*{EndFor}
\algtext*{EndProcedure}

\newcommand{\boldhdr}[1]{\textbf{#1.}}
\newcommand{\Tech}{HAGO}

\usepackage[nodisplayskipstretch]{setspace}
\newcommand*{\affaddr}[1]{#1} 
\newcommand*{\affmark}[1][*]{\textsuperscript{#1}}

\usepackage[pagebackref=true,breaklinks=true,colorlinks,bookmarks=false]{hyperref}

\def\cvprPaperID{****} 
\def\confYear{CVPR 2021}

\begin{document}

\title{Automated Backend-Aware Post-Training Quantization}

\author{
Ziheng Jiang\affmark[1]\affmark[2], \space Animesh Jain\affmark[3], \space Andrew Liu\affmark[1], \space Josh Fromm\affmark[1]\affmark[2], \space Chengqian Ma\affmark[5]\\
Tianqi Chen\affmark[2]\affmark[4], \space Luis Ceze\affmark[1]\affmark[2]\\
\affaddr{\affmark[1]Paul G. Allen School of Computer Science \& Engineering, University of Washington}\\
\affaddr{\affmark[2]OctoML}, \affaddr{\affmark[3]Amazon Web Services}, \affaddr{\affmark[4]Carnegie Mellon University}\\
\affaddr{\affmark[5]Department of Electrical \& Computer Engineering, University of Washington}\\
}

\maketitle

\begin{abstract}
   Quantization is a key technique to reduce the resource requirement and improve the performance of neural network deployment. However, different hardware backends such as x86 CPU, NVIDIA GPU, ARM CPU, and accelerators may demand different implementations for quantized networks. This diversity calls for specialized post-training quantization pipelines to built for each hardware target, an engineering effort that is often too large for developers to keep up with. We tackle this problem with an automated post-training quantization framework called \Tech{}. \Tech{} provides a set of general quantization graph transformations based on a user-defined hardware specification and implements a search mechanism to find the optimal quantization strategy while satisfying hardware constraints for any model. We observe that \Tech{} achieves speedups of 2.09x, 1.97x, and 2.48x on Intel Xeon Cascade Lake CPUs, NVIDIA Tesla T4 GPUs, ARM Cortex-A CPUs on Raspberry Pi4 relative to full precision respectively, while maintaining the highest reported post-training quantization accuracy in each case.

\end{abstract}

\section{Introduction}

Deep neural networks require an infamously large amount of computing resources (e.g., processing, memory, bandwidth). For example, the popular NLP model, BERT \cite{Devlin2018}, has layers with millions of parameters and involves a commensurate amount of compute for each inference. Although some architectures are designed explicitly with cost-reductions in mind (e.g. EfficientNets \cite{Tan2019}), they often still require tens to hundreds of megabytes to store their parameters. These requirements impose difficulties for deploying neural network models on resource-limited devices, such as edge devices (mobile, IoT, etc.), and limits deep learning uses in domains like robotics or virtual assistants where the cost of deployment is difficult to justify.

Model compression schemes can reduce the memory footprint of over-parameterized models. Pruning \cite{Cun1990} and distillation \cite{Hinton2015} remove parameters by reducing the number of network weights. Lowering precision is complementary to those techniques and is being actively explored reduce the resource requirements of neural networks without significantly harming accuracy. With hardware support, low precision training and inference can improve performance (more compute operations per second on the same hardware), reduce memory bandwidth and power consumption, and allow larger networks to fit onto a device.

However, in  real-world scenarios, there is a lack of unified toolchain that can quantize and deploy a model on a variety of hardware devices. Different hardware platforms have their own quantization toolchain, e.g., NVIDIA uses TensorRT to quantize a model for GPUs as their backend libraries can only support symmetric quantization~\cite{trt}. Similarly, Intel x86 machines rely on MKLDNN for quantization and code generation~\cite{mxnet-mkl}. Essentially, the hardware information is tightly baked in the quantization mechanism, preventing one quantization infrastructure to be applicable for different types of devices.

Additionally, different hardware devices have varying level of support for integer computation. Recently, Intel x86 machines added Intel VNNI instructions to speedup $int8$ dot product computation, where it requires the input datatype to be $uint8$ and weight data type to be $int8$ with $int32$ accumulation~\cite{mxnet-mkl}. However, ARM Raspberry Pi devices do not have any instruction that can speedup $(u)int8$ computation with $int32$ datatype. Interestingly, it has \texttt{vmlal} instruction that speeds up $int8$ x $int8$ with $int16$ accumulation or $int16$ x $int16$ with $int32$ accumulation.

In this paper, we design one toolchain that can quantize and deploy the models for many hardware platforms. In additional to versatility, our other major focus is to ensure that we can use different types of instructions available on the hardware. For example, ARM v8 \texttt{vmlal} instruction supports speeding dot product for $int8$ x $int8$ inputs and $int16$ accumulation. However, to the best of our knowledge, none of the existing toolchains can use this instruction to achieve good accuracy. For $int8$ input datatype, $int16$ datatype is not enough for accumulation, and it frequently overflows and leads to accuracy losses.

We observe that though the above observation is true, it stems from the traditional view of using all the bits in the integer datatype to represent the floating point numbers. We observe that if we differentiate between the integer data type, like $int8$, and the \textbf{effective bit width} in that datatype, like using only 6 bits instead of 8 in $int8$, we can have much higher control on the accumulation error. For example, we observe that using 6 bits in $int8$ is good enough to preserve the accuracy with $int16$ datatype. This decoupling of datatype representation and effective bit width allows us to preserve accuracy, while also using the fast instructions available on the hardware devices.

With this insight, we introduce our end-to-end system - \Tech{} - that automatically quantizes and generates high performance machine code for a variety of hardware platforms. One can denote the applicable data types for the quantized operators using a simple specification for hardware. Using this information, \Tech{} builds a search space of applicable data types and the effective bit widths for each edge of the graph. This search space is quite large. Therefore, \Tech{} performs a greedy search algorithm to find suitable effective bit lengths. Next, \Tech{} uses a deep learning compiler, like Apache TVM, to generate high performance machine code for the target platform. Specifically, this paper offers the following contributions to the greater machine learning community:

\begin{figure}[h]
\label{traditional}
\centering
\includegraphics[width=0.45 \textwidth]{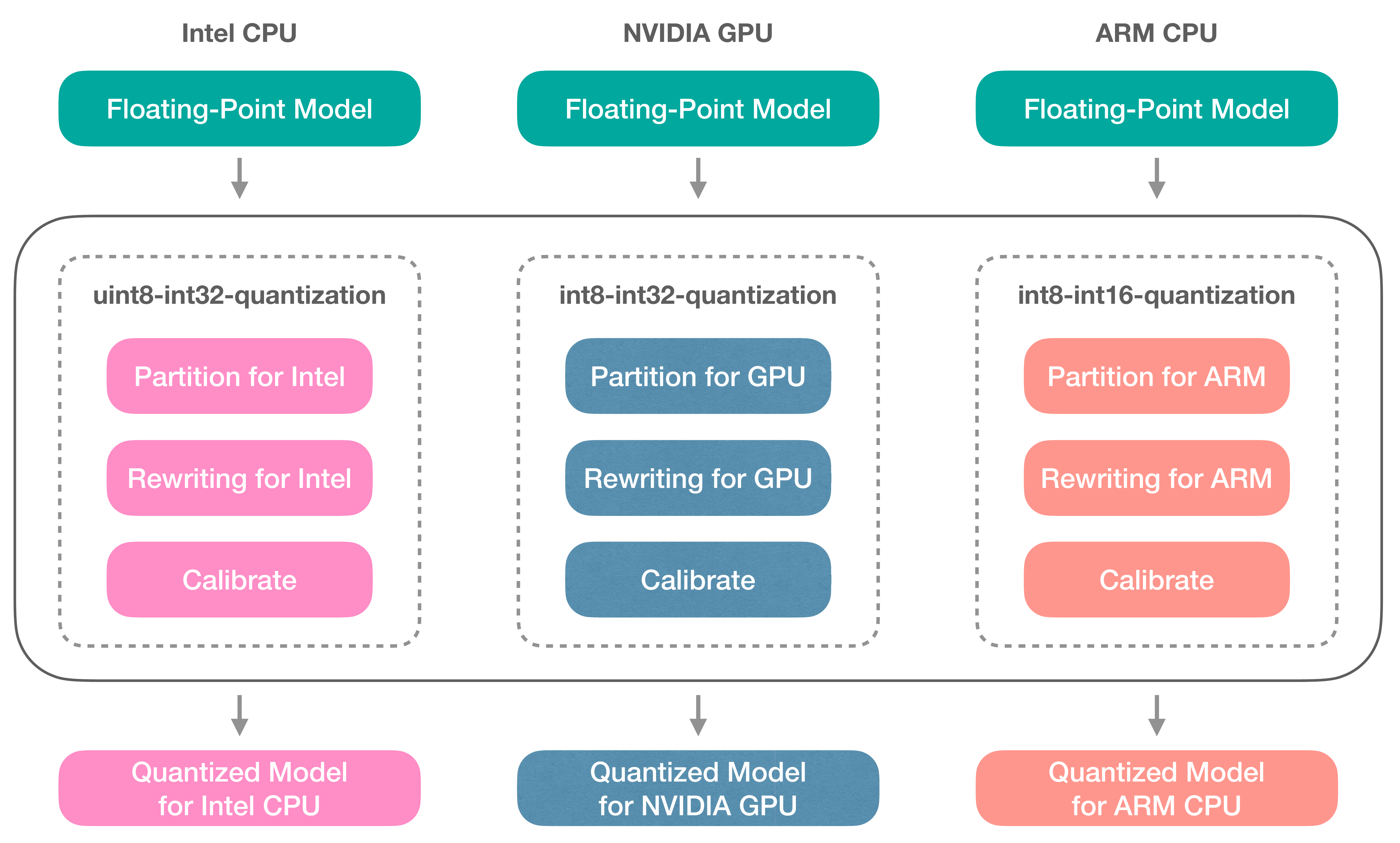}
\caption{Previous quantization framework.}
\label{fig:traditional}
\end{figure}

\begin{figure}[h]
\label{autoq}
\centering
\includegraphics[width=0.45 \textwidth]{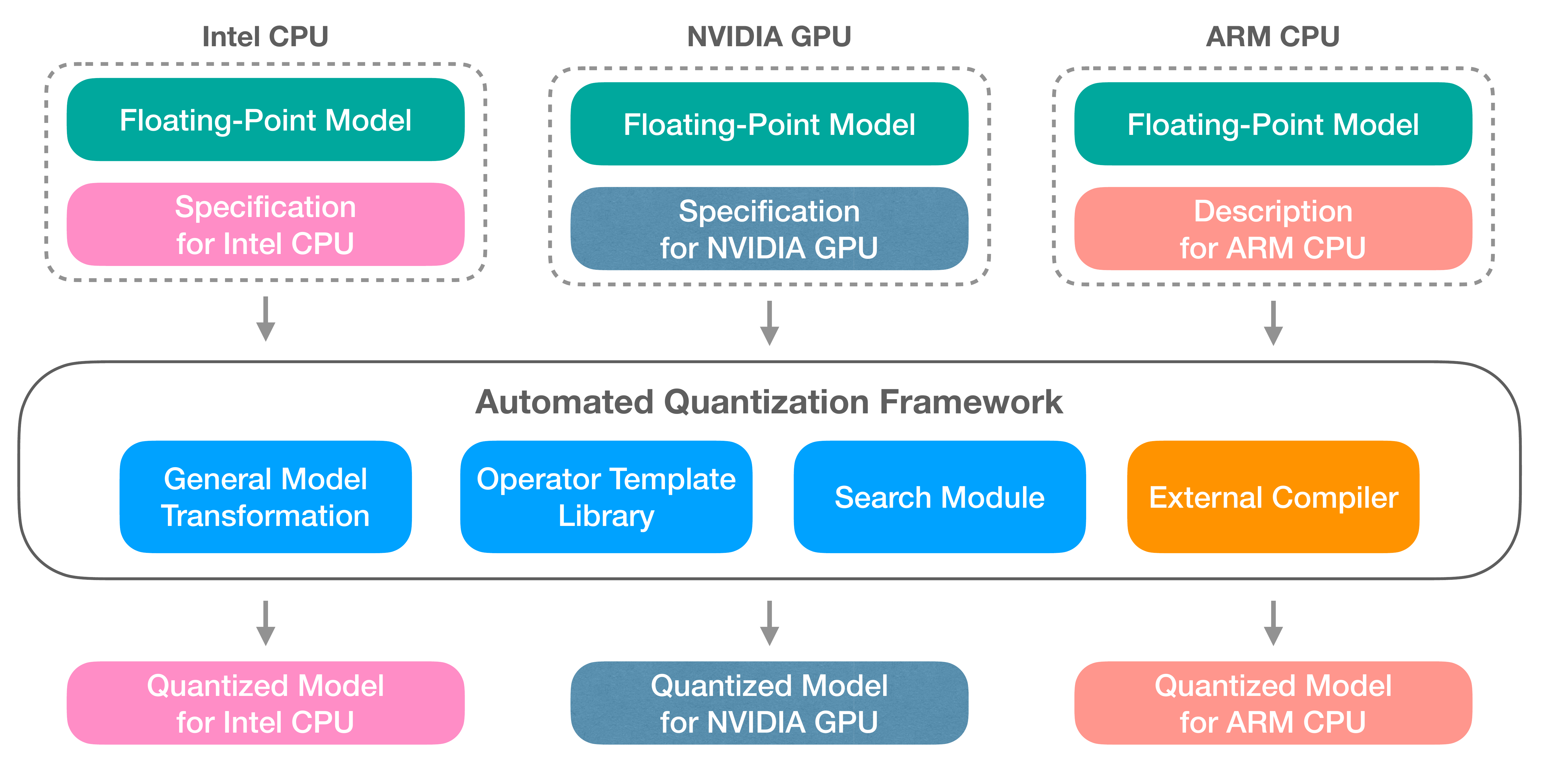}
\caption{Our automated quantization framework.}
\label{fig:autoq}
\end{figure}

\begin{itemize}
\itemsep0em 
\item We present the first end-to-end automated quantization framework, which supports configurable bitwidth quantization for many distinct hardware targets;
\item Our system leverages a novel quantization algorithm that explores optimal bitwidth quantization for each layer while meeting hardware constraints;
\item We perform extensive analysis of our system and demonstrate that it achieves state-of-the art of accuracy and performance for multiple hardware targets;
\end{itemize}

We observe that \Tech{} achieves speedups of 2.09x, 1.97x, and 2.48x on Intel Xeon Cascade Lake CPUs, NVIDIA Tesla T4  GPUs, ARM Cortex-A CPUs on Raspberry Pi4 relative to full precision respectively, while maintaining the highest reported post-training quantization accuracy in each case.

\section{Background}

There are some fundamental notations and specifications applied for this research. Generally, the goal of quantization is to convert a model running with floating-point numbers (real value) to a model running with integer numbers (quant value), without sacrificing too much accuracy. When quantizing a real value, the goal is to determine how to reduce the real value to a quant value while minimizing loss in representational fidelity. A quantized value can be viewed as $2^{bit}$ bins, where $bit$ is the number of bits in the integer type. Each of these bins typically implicitly maps to a real value. The most common and computation friendly reduction technique is linear transformation: 

\begin{equation}
s = threshold / 2^{bit - sign}
\end{equation}
$s$ represents the scale between real value and quant value and $bit$ is the number of bits we will use to represent the real values. Notably, $bit$ does not have to be the bitwidth of the integer data type exactly. Decoupling the bits of real data type and the bits used is important to compress some operators' outputs into a smaller range. For example, if we want to utilize $int16$ outputs for matrix multiplication, it is possible to set $bit=6$, which compresses the input into 6 bits and prevents overflow during accumulation even if this $int6$ value will be stored as $int8$ in most hardware targets. $sign$ represents the sign bit where $sign = 0$ implies asymmetric quantization is used, while $sign = 1$ implies the use of symmetric quantization. $threshold$ is used to approximate the maximum value in the real valued tensor. There are many methods for estimating threshold which yield varying quantization accuracy(more discussions in Section~\ref{subsec:caliberation}). For symmetric quantization, the simplest way to calculate the threshold of tensor $\textbf{A}$ is:

\begin{equation}
    threshold = max(abs(\textbf{A}))
\end{equation}

Using an estimated $threshold$, the scale $s$ is calculated and quantization can be applied using the following formula, where $r$ represents the real value of the tensor and $q$ is the value after quantizing:

\begin{equation}
q = round(r / s), r \approx q * s 
\end{equation}

\section{Overview}

\begin{figure}[t]
 \centering
  \includegraphics[width=0.5 \textwidth]{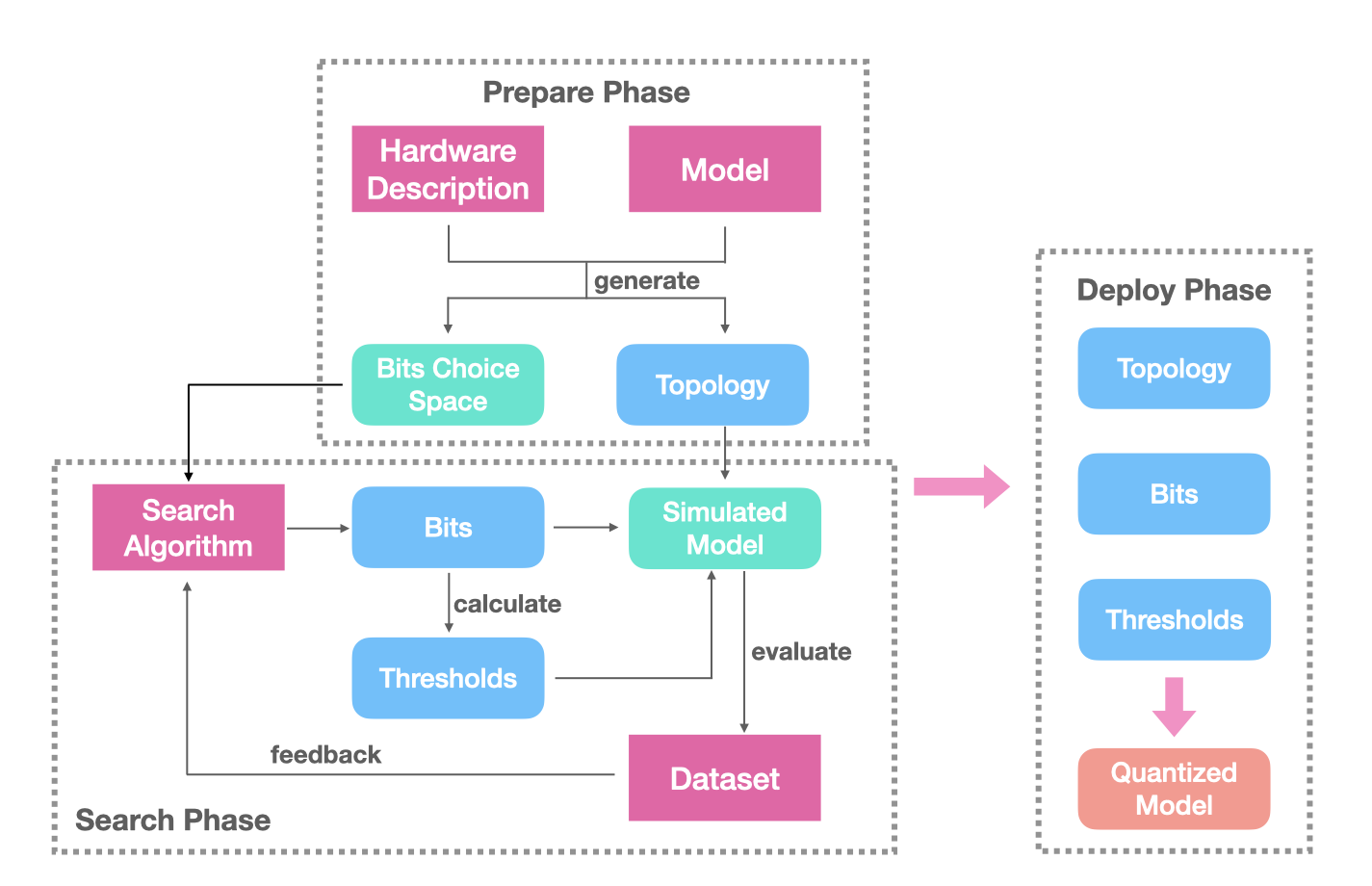}
  \caption{System overview of our quantization framework.}
  \label{arch}
\end{figure}

We present an overview of \Tech{} in Figure~\ref{arch}. \Tech{} consists of three key phases.

\noindent \boldhdr{Quantization Graph Topology Generation} In this phase, \Tech{} uses the user-supplied hardware specification to annotate which operators in the graph should be quantized. The hardware specification denotes the applicable data types for the hardware platform. \Tech{} traverses each edge in the graph representation of the real valued model and stores the specified data-types in an internal structure called the quantization \emph{Topology}. The topology is used to generate a \emph{simulated quantized graph}, where \Tech{} inserts a \texttt{simulate\_quantize} operator (described later) on each edge of the original $fp32$ graph. This operator simulates quantization error with $fp32$ datatype.


\noindent \boldhdr{Search-based Quantization} Next, \Tech{} uses a calibration dataset to find suitable ranges (or thresholds) for each tensor in the original $fp32$ model. These thresholds are used to determine quantization parameters - like scale and zero point. Meanwhile, \Tech{} analyses the topology and builds a search space for the applicable data type and effective bit widths for each quantized operator, e.g, there are 4 choices for effective bit lengths for an $int8$ input - 4 to 8. As the search space grows exponentially, any exhaustive search is intractable. We therefore use a greedy search to find the minimum bit length edge-by-edge to come up with suitable quantization parameters that preserve model accuracy.



\noindent \boldhdr{Hardware-optimized Code Generation} Finally, \Tech{} takes the resulting quantization parameters from the search phase, converts the simulated quantized model to an integer quantized model. The integer quantized model uses integer datatypes instead of $fp32$ datatype in simulated model. Next, \Tech{} uses a deep learning compiler infrastructure to generate high-performance machine code. The deep learning compiler allows code generation for a variety of hardware platforms, with the flexibility to use the low precision hardware intrinsic to achieve speedup.

In this manner, \Tech{} presents an end-to-end system capable of automatically quantizing a model and perform hardware-optimized code generation for a variety of hardware platforms.


\input{topology}

\input{search}

\input{codegen}

\begin{table}[!t]
\caption{Top1 Accuracy on the ImageNet Validation Dataset}
\label{table:acc}
\centering
\begin{tabular}{||c|c|c|c||}
\hline
\hline
Model & FP32  (\%) & I8-I32 (\%) & Drop (\%) \\
\hline
\hline
ResNet18 v1 & 70.78 & 70.52 & \textbf{0.26} \\
\hline
ResNet34 v1 & 74.41 & 73.89 & \textbf{0.52} \\
\hline
ResNet50 v1 & 76.40 & 76.32 & \textbf{0.08} \\
\hline 
SqueezeNet1.1 & 56.97 & 56.59 & \textbf{0.38} \\
\hline
MobileNet v2 & 71.24 & 69.53 & \textbf{1.71} \\
\hline 
Inception v3 & 76.71 & 76.84 & \textbf{-0.13} \\
\hline 
VGG16 & 73.11 & 72.86 & \textbf{0.25} \\
\hline
DenseNet161 & 77.48 & 77.37 & \textbf{0.11} \\
\hline
\hline
\end{tabular}
\end{table}

\section{Related Work}

Quantized-aware training~\cite{Courbariaux2016} or fine-tuning~\cite{Han2016} is one way to obtain quantized models. 
While quantized-aware training benefit from the ability to get low precision models more accurately, training is usually time-consuming and 
requires access to the full dataset, which is not always available
for optimization services. Our approach complements the existing
works on quantize-aware training on the settings where the full dataset
is not readily available.

This paper focuses on post-training quantization. Previous works
on post-training quantization~\cite{Goncharenko2018, Jacob_2018_CVPR, Choukroun2019, Meller2019} focus on building generic quantized models.
\Tech{} introduces a novel pipeline that generates optimized
quantized models based on the backend specifications.

Machine learning compilers~\cite{Chen2018a, mlir, glow} provides more flexible capabilities to generate optimized code for different quantization settings. Unlike traditional quantization
methods that focus on a single quantization scheme, \Tech{} can take full benefit of
the ML compilers to generate specifically optimized kernels for different backends.



\section{Evaluation}

\subsection{Methodology}

We evaluate \Tech{} on common vision models ResNet\cite{He2016}, Inception v3\cite{Szegedy2016}, VGG\cite{Simonyan2015}, MobileNet\cite{Sandler2018}. We implement \Tech{} on top of open-source Apache TVM (version 0.7). We evaluate \Tech{} on two server platforms on Amazon EC2 - Intel 24-core Xeon Cascade Lake CPU equipped at AWS EC2 C5.12xlarge instance and NVIDIA T4 GPU at AWS EC2 G4.xlarge instance, and the popular edge device - Raspberry Pi4 (out-of-order ARM Cortex A72). 

\subsection{Accuracy Evaluation}
First,we evaluate the effectiveness of \Tech{} in retaining the accuracy of the quantized model compared to the original $fp32$ model. We measure the accuracy across 50k images from the ImageNet validation dataset \cite{Krizhevsky2017} and show the finding in the table \ref{table:acc}. 

We observe that \Tech{} achieves less than one percent accuracy drop on all models except MobileNet. The MobileNet is hard to quantize, because batch normalization with strict limits on activation ranges (ReLu6) make the folded weights have a large dynamic range. To reduce the accuracy drop for MobileNet, quantization aware training is often required for fine-tuning the parameters.

\subsection{Performance Evaluation}

\Tech{} is designed to enable efficient deployment of quantized model across a variety of hardware platforms. In this subsection, we evaluate the effectiveness of \Tech{} in performance speedup on Intel 24-core Xeon Cascade Lake CPU equipped at AWS EC2 C5.12xlarge instance and NVIDIA T4 GPU at AWS EC2 G4.xlarge instance. Both processors have hardware support for speeding up int8 computations - Intel VNNI and NVIDIA DP4A instructions. We also evaluate several small models on the Raspberry Pi4, which do not have any fast int8 computation instructions. 

We compare the performance of the original models (referred to as TVM FP32) and the quantized models produced by \Tech{} (referred to as \Tech{} I8-I32). We execute each compiled model for 2000 images (batch size 1) and measure the average end-to-end latency. Auto-tuning \cite{Chen2018b} technique is applied to ensure high performance for both original and quantized models. 

\begin{figure}[h]
\centering
\includegraphics[width=0.45 \textwidth]{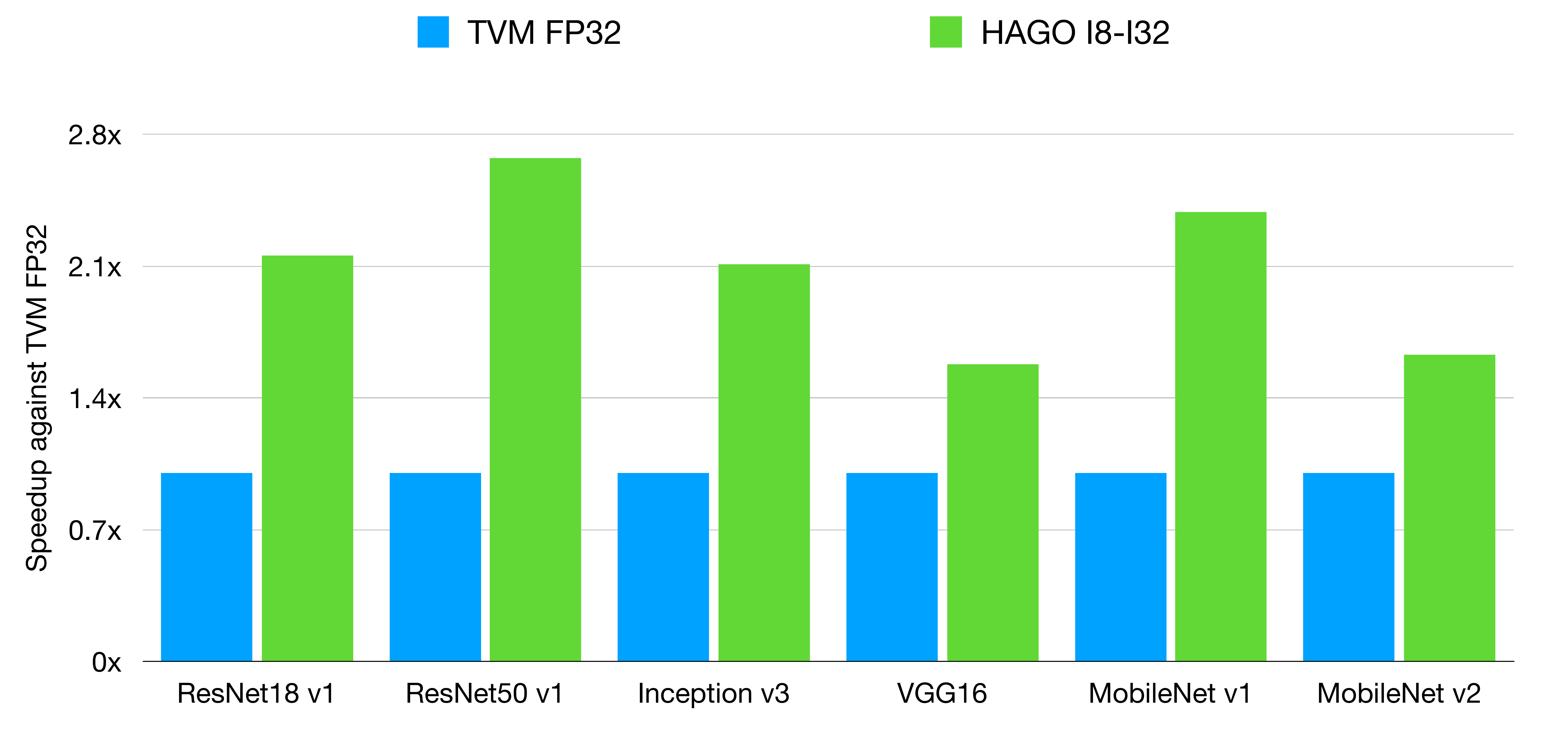}
\caption{\Tech{} Performance on Intel Cascade Lake CPU.}
\label{fig:perf_x86}
\end{figure}

\begin{figure}[h]
\centering
\includegraphics[width=0.45 \textwidth]{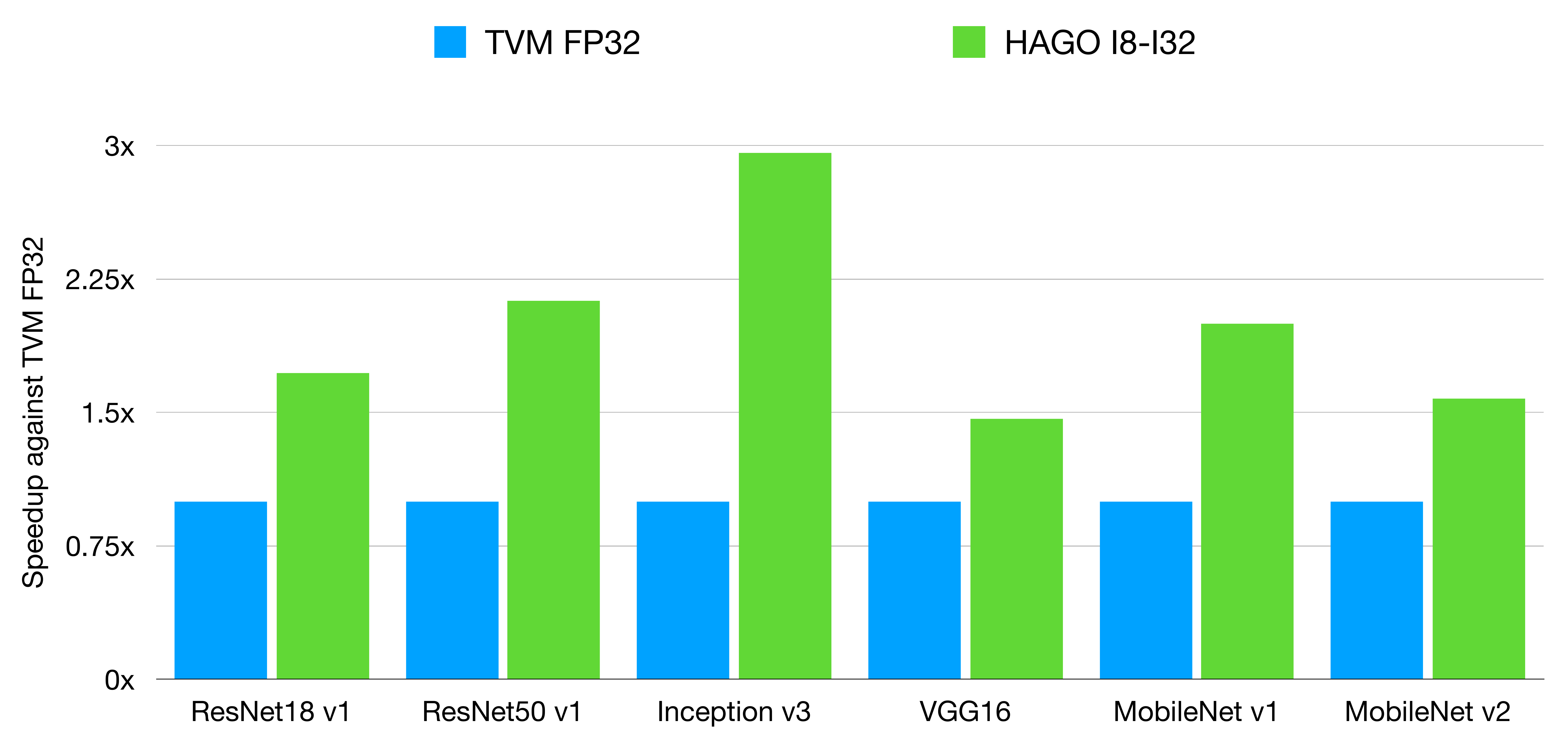}
\caption{\Tech{} Performance on NVIDIA T4 GPU.}
\label{fig:perf_gpu}
\end{figure}

\begin{figure}[h]
\centering
\includegraphics[width=0.45 \textwidth]{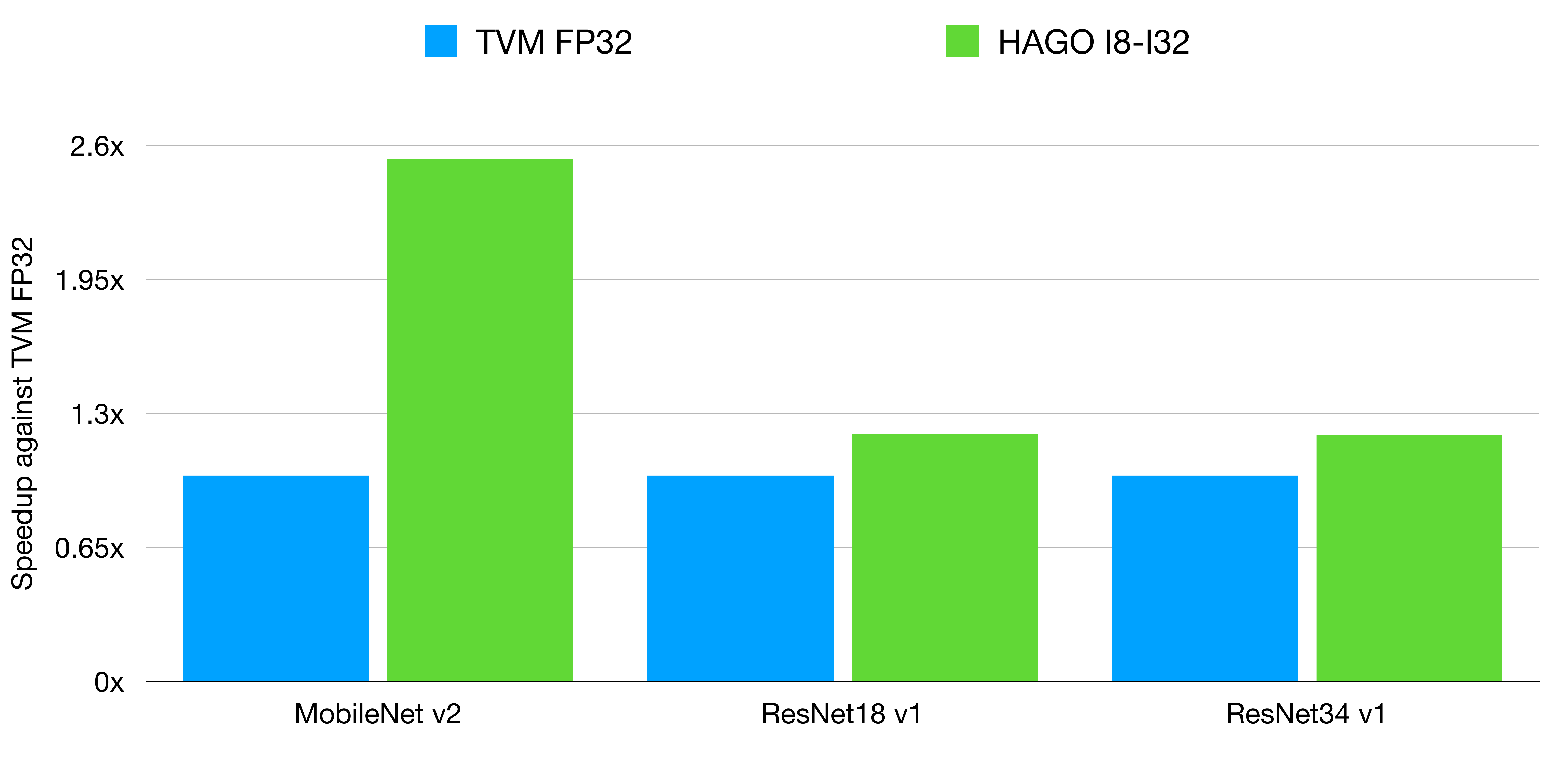}
\caption{\Tech{} Performance on Raspberry Pi4.}
\label{fig:perf_arm}
\end{figure}

As shown in \ref{fig:perf_x86}, We observe that \Tech{} achieves an average speedup of 2.09x and 1.97x for Intel Cascade Lake CPU and NVIDIA T4 GPU respectively compared to TVM FP32. For ARM CPU, we note that \Tech{} did not gain much speedup on ResNet because lack of fast int8 instruction, but it accelerate the MobileNet v2 by 2.53x. This is because that TVM stack currently lacks good depthwise convolution schedules (kernel implementation). The MobileNet is more memory bottleneck, and quantization helps on this. This shows \Tech{}'s effectiveness in providing an end-to-end solution for deploying quantized model, and supports a wide variety of hardware platforms.

\subsection{Search Effectiveness Evaluation} \label{search_eval}

In contrast to the server platforms, Raspberry Pi device do not have any $int8 \rightarrow int32$ fast instruction. However, they have vector multiply-accumulate instruction ($vmlal$) whose accumulation bit width is 16-bit that leads to better data packing in registers. In this subsection, we utilize the 16-bit accumulation instruction to generate efficient quantized model, which demonstrates \Tech{}'s effectiveness in searching good accuracy strategy under special hardware constraint.

\begin{figure}[h]
\centering
\includegraphics[width=0.45 \textwidth]{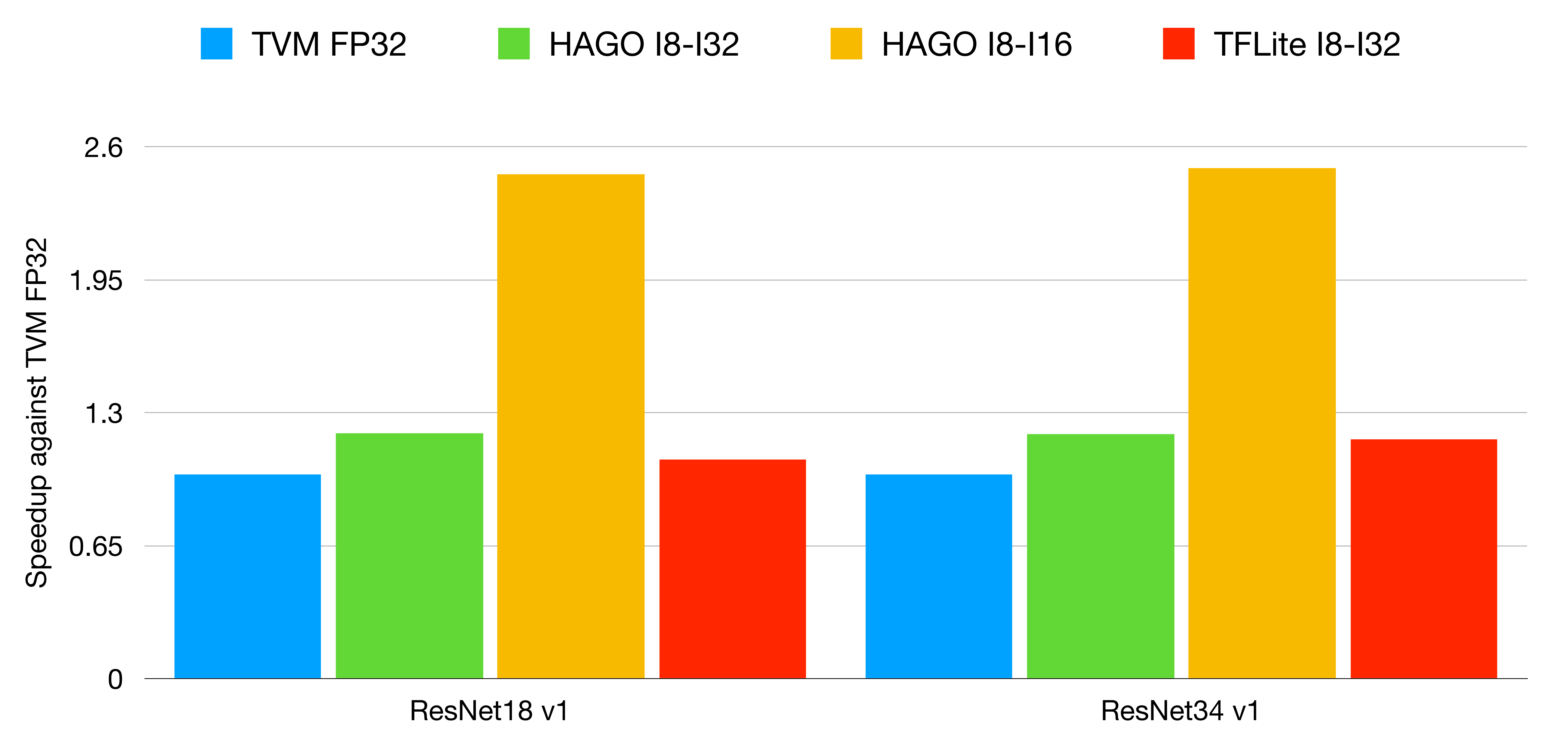}
\caption{\Tech{} Performance on Raspberry Pi4.}
\vspace{-5mm}
\label{fig:perf_int16}

\end{figure}

To utilize ARM's $vmlal$ instruction, what we need to is just to change $conv2d$'s output datatype from $int32$ to $int16$ in our hardware specification and execute the search loop to find a good accuracy setting. We experiment this pipeline with small networks like resnet18\_v1 and resnet34\_v1.  As shown in table \ref{table:int16_acc}, the greedy method achieves only 1\% accuracy drop on both $int16$ networks. We also evaluate the performance of our $int16$ models on Raspberry Pi4. The int8-int16 model perform 2.48x faster than fp32 model and 2.22x faster than tflite's int32 accumulation model, as shown in figure \ref{fig:perf_int16}.

\begin{figure}[h]
\centering
\includegraphics[width=0.45 \textwidth]{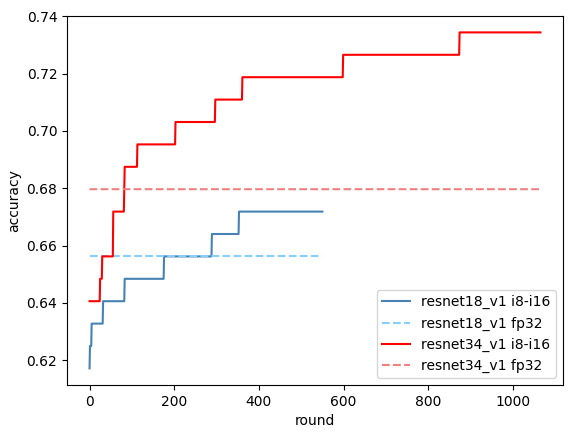}
\caption{Accuracy of int16 models on calibration dataset during the search procedure.}
\label{fig:search}
\end{figure}

\begin{table}[h]
\label{table:int16_acc}
\centering
\begin{tabular}{||c|c|c|c||}
\hline
Model & FP32 (\%) & I8-I32 (\%) & I8-I16 (\%) \\
\hline
ResNet18 v1 & 70.78 & 70.52 & 69.75 \\
\hline
ResNet34 v1 & 74.41 & 73.89 & 73.39 \\
\hline
\end{tabular}
\caption{Top1 Accuracy on the ImageNet Validation Dataset}
\end{table}

\section{Conclusion}

In this paper, we propose an automated quantization framework to the challenge of deploying efficient quantized models across a variety of hardware. While divergence existed across hardware like x86 CPU, NVIDIA GPU, ARM CPU and accelerators, we note that developers spend many effort to build specialized quantization pipeline for different hardware to gain performance speedup. This is due to the tight coupling of quantization strategy and back-end hardware property. We tackle the problem with a automated post-training quantization framework called \Tech{}. \Tech{} provides a set of general quantization graph transformation based on a user-defined hardware specification, and a set of search mechanism to find the quantization strategy meeting hardware constraints for different models. We observe that \Tech{} achieves speedups of 2.09x, 1.97x and 2.48x on Intel Xeon Cascade Lake CPUs, NVIDIA Tesla T4 GPUs, ARM Cortex-A CPUs on Raspberry Pi4 relative to $fp32$ execution, while maintaining the highest reported post-training quantization accuracy in each case.

{\small
\bibliographystyle{ieee_fullname}
\bibliography{egbib}
}

\end{document}

%% file: topology.tex
\section{Quantization Graph Topology Generation}

\subsection{Hardware Specification}

\begin{figure*}[b]
\centering
\includegraphics[width=0.80 \textwidth]{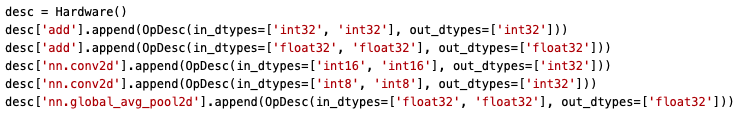}
\caption{A code snippet showing how to declare hardware's property.}
\label{fig:hardware}
\end{figure*}

We observe that different hardware devices have different integer datatype restrictions, e.g., Intel fast integer VNNI instruction prefers the input datatype to be $uint8$ and weights datatype to be $int8$ with $int32$ for accumulation, whereas ARMv8 prefers $int16$ x $int16$ data types for the input with $int32$ for accumulation. Additionally, deep learning hardware accelerators can have non-traditional requirements like using 24 bits for accumulation.

We design \Tech{} keeping this diversity in mind. \Tech{}'s objective is to have one infrastructure that can support different hardware and quantization schemes. To solve the hardware diversity problem, \Tech{} accepts a hardware specification where users users only need to define the applicable data types for each operator with a declaration style, without the need to understand the quantization logic. This differentiates \Tech{} from the traditional framework quantizers where hardware specification is tightly coupled with the quantization logic.

Currently, the hardware specification supports specific input and output data types of every operator. We present an example in Figure~\ref{fig:hardware}. This example hardware supports add operator with both $fp32$ and $int32$ data types, conv2d operator with $int16$ x $int16$ inputs with $int32$ accumulation and also $int8$ x $int8$ inputs with $int16$ accumulation, and global\_avg\_pool2d with $fp32$ datatype. Our framework will be aware of this and automatically generate quantization strategy to satisfy those constraints. 

\Tech{} reads the hardware specification and captures which operators are quantized. This can be easily achieved with the algorithm \ref{algo1}.

\begin{algorithm}[h]
\caption{Topology Generation}
\label{algo1} 
\begin{algorithmic}
\Require data-flow graph $\mathcal{G}(V, E)$, hardware specification $\mathcal{H}$
\Ensure  quantized vertices set $S_{qv}$, not quantized vertices set $S_{nqv}$
\For{$v$ in $V$ with DFS order}
\If{$v$ only support float-point computation}
\State add $v$ into $S_{nqv}$
\ElsIf{$v$ only support integer computation}
\State add $v$ into $S_{qv}$
\Else
\If{$v_i \in S_{nqv}, \forall v_i \in \mathcal{G}.src(v) $}
\State add $v$ into $S_{nqv}$
\Else
\State add $v$ into $S_{qv}$
\EndIf
\EndIf
\EndFor
\end{algorithmic}
\end{algorithm}

\subsection{Simulated Quantize Operator}
\Tech's goal is to automatically find suitable data types and quantization parameters for a given hardware specification. To allow the search, \Tech{} transform the original $fp32$ graph and inserts a new operator - \emph{simulated\_quantize} - on all the edges between the operators. Note that this is not a quantized graph. The quantization parameters like scale are not set yet. Instead, this graph is used along with hardware specification to find suitable parameter values for the simulated\_quantize op.

The simulated\_quantize operator aims at simulating the errors during quantization with fp32. It accepts scale, zero point and datatype for the input and output, and simulates the quantization error for an input tensor in $fp32$ datatype. At runtime, we can feed different values for scale and zero point, and can quickly measure the accuracy impact due to this quantization error.

\begin{figure}[h]
\centering
\includegraphics[width=0.5 \textwidth]{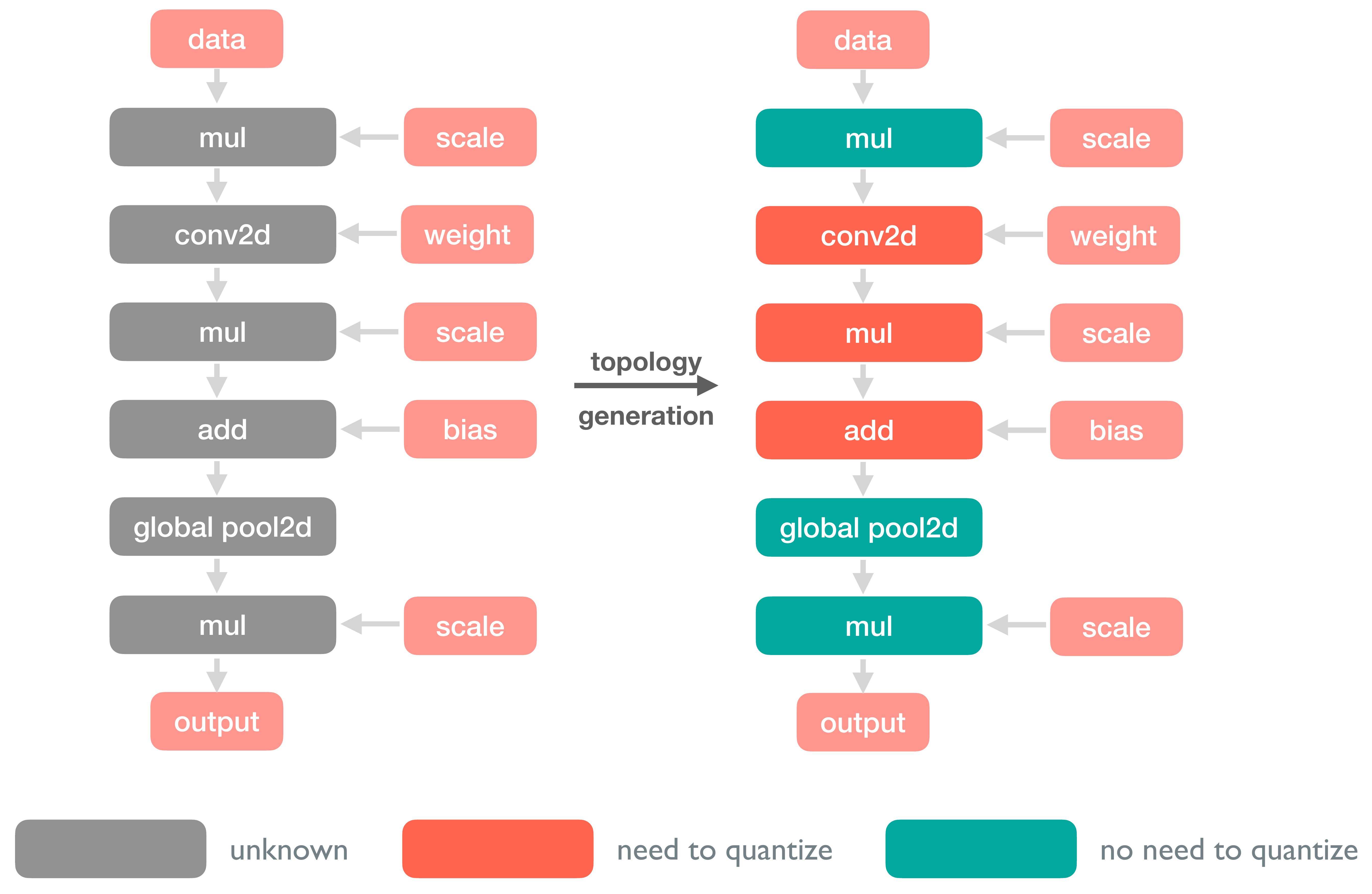}
\caption{Topology generation.}
\label{fig:topology}
\end{figure}

An example of this transformation is shown in the Figure~\ref{fig:topology}. \Tech{} first annotates which operators are quantized as per the hardware specification and creates a Topology. Next, it inserts simulated quantize operator on all of the relevant edges of the graph, e,g, inputs and outputs of conv2d and add. We also ensure that wherever possible we set the data types, e.g, the output data type for simulated quantize operator before global\_avg\_pool2d is set to $fp32$ as instructed by the hardware specification. For conv2d operator, our search based optimization will choose the right data types and number of bits.

The topology and the simulated quantized graph are then passed on to next phase of search-based optimization. At this time, we can also use topology to partition the quantized segments of the graph if, for example, a hardware accelerator supports integer only computation, allowing float computation to be offloaded to CPU.

%% file: search.tex
\section{Search-based Optimization}
We need to now identify suitable scale, zero point and data types for the simulated quantized graph. To identify these parameters, we need a range of floating point values that the integer values should represent. This range is obtained by running the original $fp32$ model on a small representative calibration dataset, and then saving the statistics of tensor values on each edge of the graph. Here, \Tech{} allows user to define or choose from existing calibration schemes to choose the range (also known as \emph{thresholds}) of tensor values. 

Next step is to use these thresholds to find the minimum number of bits for each edge in the graph. To do this, \Tech{} first analyzes the topology to build a search space that represents all the possible number of bits for each edge in the graph. Then we employ a feedback-driven search to wisely navigate this search space. At each point in the search space, we use Model Simulation, i.e., we use $simulated\_quantize$ operators that simulate the quantization with $fp32$ to measure the goodness of a configuration point. At the end of this search, we end up with suitable scale and zero points for the simulated quantized model.

\subsection{Statistics Collection}
\label{subsec:caliberation}
As mentioned above, we use the original $fp32$ model and a small calibration dataset to collect statistics of all the tensors in the graph. Then, we use a calibration scheme to estimate the threshold. For determining the threshold from the collected outputs, the following strategies can be used:

\begin{itemize}
\itemsep0em 
\item \textbf{Max range}: use the maximum value of the output as the threshold of the corresponding edge.
\item \textbf{Average range}: use the quantile of range as the threshold, so that we can remove some outliers.
\item \textbf{KL estimate}: choose a threshold that makes the KL distance between real output and quantized output small enough.
\end{itemize}

In our experiments, KL estimate gives us better accuracy, but it can be quite time consuming, making it not the best approach especially if we want to use search. Also, we observe that average range always leads to better accuracy than the max range method. Here, a user can also define their calibration technique to fit one's hardware requirements.

We also provide an option to round the calculated threshold to the nearest power-of-two number. This make it possible to use shifting to replace multiplication and gives better performance in the final quantized model, also enable applications on some accelerators which do not have integer multiplier.

\subsection{Model Simulation}
We use simulated quantize to simulate the errors that happen due to quantization. The error can come from several aspects: First, the \textbf{rounding error} happens when round float-point number to integer. Secondly, the \textbf{saturated error} can happen when the casted integer is clipped to suite the range. Last, there is also the \textbf{overflow error}, possible when the output of quantized operator is beyond the limit of output data type can represent.

We simulate these errors in $simualted\_quantize$ operator using $fp32$ computation. This allows us to quickly measure the accuracy of configuration on the calibration dataset in our search phase. After all the preparation described above, now the quantization problem is converted to a search problem: the aim is to find the best setting from the search space, with the output(accuracy) on the calibration dataset every round as the feedback, for achieving the best accuracy (or other objectives like performance or memory consumption) on the simulated model.





\subsection{Search Space Generation}

\begin{figure}[h]
\centering
\includegraphics[width=0.5 \textwidth]{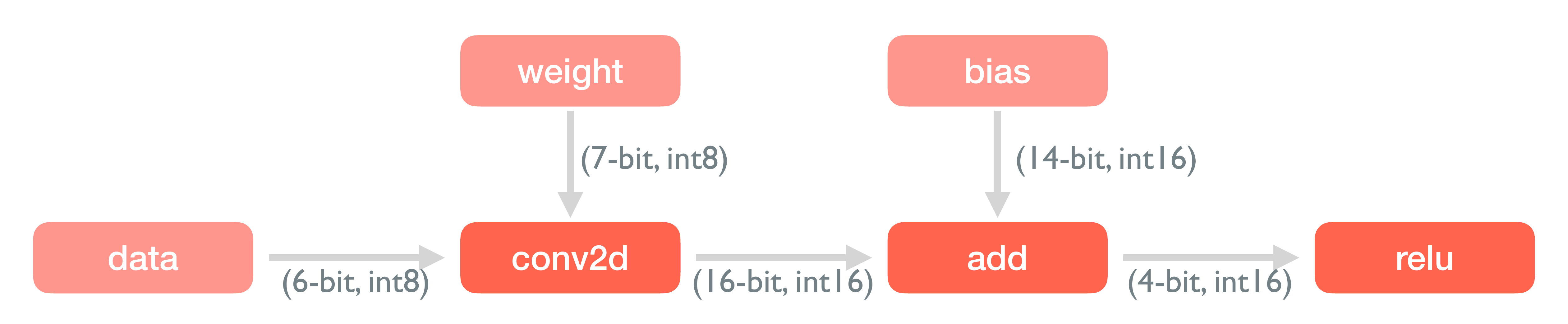}
\caption{The searched best quantization strategy for int16 resnet16 v1.}
\label{fig:strategy_case}
\end{figure}

Unlike the existing quantization framework, which calibrates the quantized model only by adjusting the threshold, we aim to find the number of used bit for every tensor under the given hardware constraint. Conventionality, the number of bit will be set as the maximum bit width the data type allows. We found that this leads to severe accuracy drop in some case like supporting $int8 \rightarrow int16$ quantization. This is because if we execute conv2d with 8-bit operands, it would be easily to get overflow for int16 activation and generate meaningless outputs. To address this issue, \Tech{} allows use only subset bit width of the real data type, and make it as search-able parameter. For example, we can compress the input into 6-bit to avoid overflow, even still stored with int8 in the final compiled model. We verified this assumption by checking the best quantization strategy on int16 accumulation ResNet18 v1 as shown in figure \ref{fig:strategy_case}.

Although we can achieve more compact compression by scaling the threshold, choosing the bit width as parameter brings several advantages: 1. easily to generate search space according to the hardware specification; 2. a explicit bit width information can help us decide the final data type we would like to use after lowering; 3. the final strategy with bit width on every tensor is easier to understand and interpret than thresholds.

Now the goal is to construct the search space given a hardware specification and the original model. The allowed data type on this edge decides the feasible bits. The bit range on this edge is from the pre-defined lowest bit width to the maximum bit width this datatype supports. the lowest bit width is a constant that was set to reduce the search space. During our experiments, it was set as 4, since a range smaller than 4 often results in meaningless findings. For example, if this edge can be int8, the bit range of this edge is from 4 to 8, , if this edge can be int16 or int32, then the bit range would be 4 to 16 or 4 to 32. 

Assuming that the bit range on every range $R_i$,  the search space can be represented as $R_0 \times R_1 ... \times R_\mathcal{E}$. Here $\mathcal{E}$ is the number of edges in the model. The search space can be pretty large. Take resnet18\_v1 as example, it has 118 edges so the search space is larger than $4^{118}$. Design the search algorithm to find workable solutions from the space is a key question.

\subsection{Search Methods}

By noticing that the set of feasible solutions is discrete, it is viable to put it a classic combinatorial optimization problem, a topic that consists of finding an optimal object from a finite set of objects. 

Since the search space is quite large, \textbf{exhaustive search} is not possible. In our preliminary experiments, we find that \textbf{random search} is not good enough to find a valid combination. It is because a single bad layer configuration can cause great damage to the final output. As a result, feasible solution points are quite sparse in the space. 

From this observation, we apply \textbf{simulated annealing} algorithm - a metaheuristic to approximate global optimization in a large search space. For this problem, we define $E = exp(-cost/T) $ as the $energy$, where the $cost$ is negative of accuracy evaluated with a candidate (effective bit vector) and $T$ is a hyper-parameter called $temperature$ in simulated annealing. By adding random disturbance to the effective bit on every edge, we can get the new energy value $E_{new}$. By compare $E_{new}$ and $E_{old}$, the new candidate can be accepted with certain probability.

Simulated annealing works good to produce some feasible solutions for our problem, but still need many rounds to "warm-up" - find a good candidate because of the sparsity. Also, there are many hyper-parameters to be set, which maybe different across tasks. So a smarter algorithm is preferred.

One assumption is, for the effective bit on a specific edge, we do not need to try the lower bit if the higher bit already brings accuracy drop. Based on the assumption, we design a \textbf{greedy search} algorithm which try to reduce the effective bit edge by edge as long as it does not hurt performance. In detail, the algorithm use the maximum bit width on every edge as the initial candidate. Starting with the first edge in the DFS order, the algorithm reduces the effective bit on this edge by one. If the accuracy drops, it moves to the next edge, otherwise keep reducing the effective bit on this edge. We also attach a formal description (algorithm \ref{algo2}) to explain the greedy search method. In section \ref{search_eval}, our experiments shows that the greedy search approach finds the feasible strategy with a limited number of iterations.

\begin{algorithm}[h]
\caption{Greedy Search Algorithm}
\label{algo2} 
\begin{algorithmic}
\Require data-flow graph $\mathcal{G}(V, E)$, the bit choice range on every edge $\mathcal{B}(e_i)$, number of rounds $\mathcal{R}$
\Ensure  the best solution (number of bits on every edge) $\langle b_0, b_1, ..., b_e \rangle$
\State initial candidate with the max bit on every edge: $b_i \gets max(\mathcal{B}(e_i))$ for $e_i \in E$ 

\For{$r=0$ \textbf{to} $\mathcal{R}$}
\For{edge $e_i$ in the DFS order}
\State{$\hat{b_i} \gets b_i - 1$ }
\State{$\mathcal{M} \gets$ generate simulated model from $\langle b_0, ... \hat{b_i}, ..., b_e \rangle$}
\State{$\mathcal{L} \gets $ evaluate $\mathcal{M}$ on $\mathcal{D}$}
\If{$\mathcal{L} < $ previous best $\mathcal{L}$}
\State{$b_i \gets \hat{b_i}$}
\Else
\State{continue}
\EndIf
\EndFor
\EndFor
\end{algorithmic}
\end{algorithm}

%% file: codegen.tex
\section{Hardware-Optimized Code Generation}
After search phase is over, we have now identified suitable scale and zero points of the quantized model. Now, we first convert this simulated quantized model ($simualte\_quantize$ works on $fp32$ data types only) to an actual quantized model that works on integer data types. This step is called Quantized Model Realization step. Next, \Tech{} employs a deep learning compiler to generate high performance machine code for the given hardware platform. Now, we present the details of the steps.

\subsection{Quantized Model Realization}
This step converts a simulated quantized model to an integer quantized model, i.e., it converts simulated quantize operators to more formal operators - quantize, dequantize and requantize - that work directly with integer data types. We also handle any adjustment due to scale or zero points. For example, mobilenet networks have RELU6 which is basically a clip operator with limits as 0 and 6. However, numbers 0 and 6 are valid only for $fp32$ numbers. We ensure that while realizing the quantized model, we convert the min and max of $clip$ operator as per the output scale of simulated quantize operator just before clip operator to maintain the validity.

\subsection{Deep Learning Compilation}
\Tech{} employs a deep learning compiler toolchain, specifically Apache  TVM~\cite{Chen2018a}, to generate high-performance code for the given hardware platform. Deep learning compilers, like TVM, are typically composed of two levels of intermediate representations - graph-level and low-level tensor representation. A model is first represented at the graph level, and is then iteratively optimized, first at the graph-level and then at the tensor-level to yield a high-performance machine code. TVM relies on stable code generators like LLVM~\cite{Lattner2004} and NVCC~\cite{wiki:nvcc} to support a wide variety of hardware platforms.

\boldhdr{Graph-level IR} TVM supports a variety of deep learning operators - like conv2d, dense, relu etc. For quantization, it has an extension called QNN~\cite{Jain2020} that supports relevant operators like quantize, dequantize and requantize. In the quantized model realization phase, \Tech{} uses QNN operators to realize a graph. The resulting graph goes through a series of optimizations, like operator fusion and dead code elimination, to produce an optimized graph.

\boldhdr{Low-level Tensor IR} Next, each graph-level operator is represented in a loop-based low-level IR to express the real computation. Here, QNN operators rely on integer-only computations as much as possible to maintain speedup. Furthermore, different hardware vendors have different instructions for speeding integer-computation. Intel x86 CPUs have VNNI, Nvidia GPUs have DP4A and ARM has VMLAL. Here, TVM allows developers to write kernels that use these fast instructions to speedup important operators, like conv2d and dense. Finally, TVM uses LLVM or NVCC to generate machine code.

\noindent In our experiments, we reuse the existing graph and tensor-level optimizations to speedup the integer quantized models on a variety of hardware devices.

\subsection{Batched GPU Search}
Even after improving the speed of quantized model with TVM, the search procedure can still take a long time. The bottleneck is in compiling the model again and again for different quantization strategies, and evaluate them on the calibration dataset. To accelerate the search, we design put the search options as into the arguments of the simulated quantize operator, so that we can compile the model once and evaluate multiple configurations simultaneously. Using this feature, we can batch the search configurations and perform \emph{batched search} on the GPUs. The batching based approach boost our overall search procedure from one day to less than two hours.